\documentclass[dvipsnames]{article}
\usepackage{arxiv, times}
\usepackage{graphicx}
\usepackage{balance}
\usepackage{booktabs}
\usepackage{blkarray} 
\usepackage{amsmath,amsfonts,amssymb}
\usepackage{amsthm}
\usepackage{mathtools}
\usepackage{algorithm}
\usepackage[noend]{algpseudocode}
\usepackage{subfigure}

\usepackage[utf8]{inputenc}

\graphicspath{{figs/}}

\newcommand{\nnz}{\mathrm{nnz}}

\newcommand{\cO}{\mathcal{O}}
\newcommand{\cC}{\mathcal{C}}
\newcommand{\cG}{\mathcal{G}}
\newcommand{\cE}{\mathcal{E}}
\newcommand{\cV}{\mathcal{V}}
\newcommand{\cP}{\mathcal{P}}

\newcommand{\papertitle}{Scaling Graph Clustering with Distributed Sketches}

\usepackage{hyperref}
\hypersetup{
    unicode=false,
    pdftoolbar=true,
    pdfmenubar=true,
    pdffitwindow=false,
    pdfstartview={FitH},
    pdftitle={\papertitle},
    pdfauthor={Benjamin W. Priest},
    pdfsubject={\papertitle},
    pdfkeywords={Sketching algorithms, Distributed Algorithms, Massive Graphs, Graph Clustering},
    pdfnewwindow=true,
    colorlinks=True,
    linkcolor=BurntOrange,
    citecolor=RawSienna,
    filecolor=magenta,
    urlcolor=blue
}
\newcommand{\email}[1]{\tt\small\href{mailto:#1}{#1}}

\title{\papertitle\thanks{This work was performed under the auspices of the U.S. Department of Energy by Lawrence Livermore National Laboratory under Contract DE-AC52-07NA27344  (LLNL-CONF-812693) and was supported by the LLNL-LDRD Program under Project No. 20-FS-037. Experiments were performed at the Livermore Computing Facility.}}

\author{
Benjamin W. Priest,\,\,\,\,\,\, Alec Dunton$^\dagger$,\,\,\,\,\,\, Geoffrey Sanders \\[0.3cm]
       {Center for Applied Scientific Computing, Lawrence Livermore National Laboratory}\\[0.2cm]
       {$^\dagger$Department of Applied Mathematics, University of Colorado Boulder}\\[0.2cm]
       \texttt{\email{priest2@llnl.gov}, \email{alec.dunton@colorado.edu}, \email{sanders29@llnl.gov}}
}

\date{}

\iclrfinalcopy

\begin{document}

\maketitle

\begin{abstract}
The unsupervised learning of community structure, in particular the partitioning vertices into clusters or communities, is a canonical and well-studied problem in 
exploratory graph analysis.
However, like most graph analyses the introduction of immense scale presents challenges to traditional methods. 
Spectral clustering in distributed memory, for example, requires hundreds of expensive bulk-synchronous communication rounds 
to compute an embedding of vertices to a few eigenvectors of a graph associated matrix.
Furthermore, the whole computation may need to be repeated if the underlying graph changes some low percentage of edge updates. 
We present a method inspired by spectral clustering where we instead use matrix sketches derived from random dimension-reducing projections. 
We show that our method produces embeddings that yield performant clustering results given a fully-dynamic stochastic block model stream using both the fast Johnson-Lindenstrauss and CountSketch transforms.
We also discuss the effects of stochastic block model parameters upon the required dimensionality of the subsequent embeddings, and show how random projections could significantly improve the performance of graph clustering in distributed memory.
\end{abstract}

\section{Introduction} \label{sec:intro}

Analysts working in applications as widely ranging as biology, sociology, network science, computer science, telecommunications, and others deal regularly with graph-expressible data where a major task of interest is to find structure, usually defined as a partitioning of ``like'' vertices into clusters.
Although many graph clustering algorithms have arisen in the literature, 
and several works make important contributions to improving their distributed versions
\citep{Que15Louvain, Halappanavar17GC, Ghosh18GC, Liu18GC},
applying them to scales that require distributed implementations on modern systems (topology data is near PetaScale, say $>$ 100B edges) remains challenging. 

We will center our focus in this paper to accelerating algorithms in the form of spectral clustering~\citep{von2007tutorial}.
In the most generic form, spectral clustering computes approximate eigenpairs of a graph-dependent matrix
(commonly adjacency, Laplacian, or a centered/normalized variant).
Then, $k$ extremal eigenvectors form an \emph{embedding} of the vertices into $\mathbb{R}^k$.
Some downstream conventional clustering algorithm (e.g. $K$-means~\citep{lloyd1982least} or dbscan~\citep{hahsler2019dbscan}) then partitions the vertices using their embedded locations.
Often recursion is employed, where the initial embedding reveals several of the {\em best} partitions ({\em strongest} communities) and reapplying 
the process on much smaller sub-partitions to further resolve community structure. 

Although spectral clustering is a popular method and reasonably performant (log-linear cost and storage in input data) for real-world serial applications, 
it is not currently employed at extreme scales.
State-of-the-art eigensolvers such as Krylov/Lanczos methods are iterative methods, 
relying on sparse linear algebraic kernels (primarily sparse matvec, dot-product, and their block vector variants), 
for which high-quality scientific computing packages are available \citep{Hernandez:2005:SSF,Anasazi}.   
Recent work develops these linear algebra kernels and others for challenging graph topology \citep{bulucc2011combinatorial}.
Distributed memory implementations of eigensolvers require hundreds of resource-hungry bulk-synchronous operations (sequences of sparse matvecs).
Numerical stopping criteria are also poorly understood for general, large real-world graphs.
Finally, conventional spectral embedding algorithms are not 
efficient in evolving graph applications, where an update of a few edges (say those connecting previously poorly connected partitions) can cause
previous solutions to take longer to iteratively improve than random initial guesses would.

In this document we present an alternative method relying upon a cheaper and more scalable embedding procedure using \emph{linear sketches}: dimension-reducing linear projections drawn from a carefully-chosen distribution.
Random sketch matrices allow us to approximately preserve row inner products and norms of an arbitrary matrix in a much lower dimension with high probability.
Our method replaces an expensive spectral embedding with an efficient - though coarse - linear sketch embedding. 
In addition to gaining computational efficiency, linear sketches are designed for the turnstile streaming data model - i.e. they are indifferent to the order in which matrix items are received, and are robust to changes in the underlying matrix. 
Thus, our method produces a vertex embedding that features (1) computation and communication linear in the number of edge updates,  (2) a simple distributed memory implementation, and (3) natural robustness to dynamic data.  

Others have broadly applied matrix sketches throughout numerical linear algebra, often with an emphasis on matrix multiplication, regression, or low-rank approximation \citep{woodruff2014sketching}.
\cite{ailon2009fast} introduced a fast formulation of the classic Johnson-Lindenstrauss transform (JLT, \citep{johnson1984extensions}) with the fast approximation of nearest neighbors in the embedding space as a motivating example.
\cite{traganitis2015sketch} devised a library of tools SkeVa that utilize sampling to produce approximate $K$-means clustering on high-dimensional data.  
Our approach is partially inspired by \cite{gilbert2012sketched}, who used JLTs to directly, though coarsely, approximate the singular values and vectors of large matrices.

Many recent graph and matrix sketching applications make use of linear $\ell_p$-sampling sketches \citep{jowhari2011tight} based upon precision sampling \citep{andoni2011streaming}.
\cite{ahn2012analyzing, ahn2012graph, ahn2013spectral} applied linear sampling sketches to estimate graph properties  including approximating spectral sparsification, which \cite{kapralov2017single} improved with a single-pass algorithm.
\cite{ahn2018hypergraph} described spectral clustering algorithms on hypergraphs, and used sampling sketches to accelerate their algorithms.
Unfortunately, $\ell_p$-sampling sketches are tricky to implement and are not considered efficient for large scale applications.

Others have utilized sketches as a means to accelerate spectral clustering.
\cite{fowlkes2004spectral} describe a method using a Nystr\"{o}m low-rank approximation to the Laplacian by way of column sampling, and later using the eigenspectrum of the Nystr\"{o}m approximation to construct an embedding to be clustered in $\cO(nmk + m^3)$ operations, where $n$ is the number of vertices, $m$ is the dimension of the Nystr\"{o}m approximation, and $k$ is the embedding dimension.
\cite{li2011time} later improved the complexity to $\cO(nmk)$.
We will describe a method that requires $\cO(\nnz(X))$ time, where $\nnz(X)$ is the number of nonzero elements in the graph matrix $X$ - i.e. twice the number of edges. 
\cite{gittens2013approximate} use a conventional power iteration method to coarsely  approximate eigenvectors before utilizing JLTs to project into a small dimensional space.
Our method avoids the expensive power iteration process.
Additionally, unlike either of these methods, our embeddings scale naturally to the distributed model, take advantage of the sparsity structure in the graph matrix, and are fully robust to dynamic and streaming graphs.

We consider popular efficient random matrix projections in our analyses:
the fast Johnson-Lindenstrauss transform based upon the Walsh-Hadamard transform (FWHT)~\citep{ailon2009fast}, and
the CountSketch transform (CST)~\citep{clarkson2017low}.
We will produce results comparing the performance of FWHT- and CST-based embeddings of stochastic block model (SBM) vertices into lower-dimension space.
In our experiments, we employ UMAP~\citep{mcinnes2018umap} to perform an additional non-linear dimensionality reduction and HDBSCAN~\citep{mcinnes2017hdbscan} to cluster the resulting embeddings.
We further present scaling results demonstrating that our linear embedding procedure can embed SBMs with tens of billions of edges in seconds on a modest compute cluster.

\section{Notations and Background} \label{sec:notations}

We will assume throughout an undirected, unweighted, unsigned, connected and static graph $\cG = (\cV, \cE)$ with vertex and edge sets of size $|\cV| = n$ and $|\cE| = m$, respectively.
Spectral clustering methods generalize trivially to weighted graphs and 
other have extended them to handle signed graphs \citep{knyazev2018spectral}, 
dynamic  graphs \citep{Ning2007}, and directed graphs \citep{VanLierde}.
The sketching methods we employ generalize to weighted, signed, dynamic, and directed graphs with much less complication, as the key operator is a simple linear projection.
In cases where signal to noise ratio is high enough, the sketching approach likely provides scalability that is not possible in the existing approaches.  
For example, some directed graph methods involve complex-valued iterative eigensolvers, which are challenging to implement efficiently at scale on graph topologies. 
This is particularly the case for sketch-based dynamic graphs \citep{martin2017fast}.

In particular, we will assume that $\cG$ is sampled from an undirected stochastic block model with $c$ communities (or blocks) with symmetric probability matrix $P \in [0, 1]^{c \times c}$ and the community size vector $\cC \in \mathbb{Z}_+^c$.
The $(i, j)$th entry of $P$ indicates the pairwise probability that each of the $\cC_i$ vertices in community $i$ are neighbors with each of the $\cC_j$ vertices in community $j$. 
In general, the diagonal entries of $P$ will be larger than the off-diagonal entries, which is meant to simulate the inter- and intra-connection densities of ground truth communities in empirical networks. 
The expected ratio between the diagonal entries of $P$ and and off-diagonal entries of $P$ are determined by additional parameters $\rho_{in}$ and $\rho_{out}$, which parameterize the distributions from which the elements of $P$ are sampled.

A graph $\cG$ has adjacency matrix $A$, where $A_{x, y} = 1$ iff $xy \in \cE$ and is zero otherwise.
We will embed the rows of the adjacency matrix because the SBMs we consider are nearly regular, and so little is gained by utilizing a Laplacian formulation.

We use {\em pythonic} notation for matrix rows and columns
(the row vector corresponding to the neighborhood of vertex $i$ is $A_{i,:}$ and the $i$-th column is $A_{:,i}$).

Canonical adjacency-based spectral embedding of dimension $k$ on $\cG$ is performed as follows:
\begin{enumerate}
  \item Let $V \in \mathbb{R}^{n \times k}$ be the matrix whose columns $V_{:, 1}, \dots, V_{:, k}$ are the eigenvector of $A$ associated with most positive eigenvalues. 
  \item Identify with each vertex $x$ the row vector $V_{x, :}$.
\end{enumerate}
We call the rows of $V$ a \emph{spectral embedding} of their corresponding vertices.
The spectral clustering algorithm consists of computing a spectral embedding and executing a clustering algorithm on the embedded vectors, i.e. the rows of $V$.

\section{Matrix Sketching} \label{sec:sketch}
Matrix sketching is a numerical linear algebraic tool with applications in, e.g., latent semantic indexing~\citep{papadimitriou2000latent}, low-rank approximation~\citep{halko2011finding}, and least-squares problems~\citep{rokhlin2008fast}. 
A primary goal of matrix sketching is to embed a matrix $X \in \mathbb{R}^{n \times p}$, comprised of $n$ data points in a $p$-dimensional feature space, in a lower dimensional space such that geometric properties of the original matrix are preserved to a desired level of fidelity. 
Mathematically, this typically entails the application of a linear operator $S \in \mathbb{R}^{p \times s}$ with $s \ll p$ to form the {\it sketch matrix} $XS \in \mathbb{R}^{n \times s}$.\footnote{Similar sketching procedures involving multiplication on the left-hand side of the argument, as well as those which embed both the row-space and column-space of $X$ (bi-linear sketches), can be defined analogously.}

For the scope of this work, we seek sketching matrices that
\begin{enumerate}
	\item preserve pairwise distances between rows in $X$ to within a tolerance which we denote $\varepsilon$, with a constant failure probability; 
	\item satisfy 1) by projecting into $\Theta(\varepsilon^{-2}\log n)$ dimensions;
	\item admit scalable and sparse distributed memory implementations.
\end{enumerate}
To this end we consider the FWHT, which satisfies conditions 1) and 2), but we will show that it struggles to satisfy our needs for condition 3).
We also consider the CST and show that it exhibits superior distributed memory scaling as compared to FWHT, although it does not provide as strict a guarantee for condition 1). 
In particular, CST is \emph{not} a Johnson-Lindenstrauss transform. 
However, it has been applied to great effect throughout the literature even with its less strict guarantee~\citep{woodruff2014sketching, clarkson2017low, yang2020reduce}.
We extensively show that CST and FWHT create embeddings of similar quality in our experiments in Section~\ref{sec:experiments}.

In rough terms, the Johnson-Lindenstrauss lemma states that there is distribution of linear operators that can embed any $n$ points in a $p$ dimensional feature space into  $\Theta(\varepsilon^{-2}\log(n))$ dimensions such that all pairwise inner products are preserved to within a factor of $(1\pm\varepsilon)$ with a constant probability of failure~\citep{johnson1984extensions}. 
That is, our embedding dimension is independent of the dimension of the feature space and logarithmically dependent on the number of points which we are embedding. 
The FWHT embeds a matrix $X$ as $XDPS \in \mathbb{R}^{n \times s}$ where $D \in \mathbb{R}^{p \times p}$ is a diagonal matrix whose entries are i.i.d. $\pm 1$ with equal probability, $P \in \mathbb{R}^{p \times p}$ is a Hadamard matrix, and $S \in \mathbb{R}^{p \times s}$ is a sparse matrix whose nonzero entries are one to indicate uniform subsampling.
In practice $p$ is assumed to be a power of two.
This construction allows the FWHT to satisfy the Johnson-Lindenstrauss lemma with asymptotically lower complexity than prior  JLT formulations.

$D$, $P$, and $S$ are never actually formed in practice, as their entries can be generated quickly as needed. 
In particular, embedding an element $X_{i,j}$ amounts to sampling column indices $k_1, \dots k_s$ and generating a vector $X_{i, j} * D_{i, i} * [H_{j, k_1}, \dots H_{j, k_s}]$.

The CST was first developed for numerical linear algebra by \cite{clarkson2017low} and was inspired by the celebrated CountSketch~\citep{charikar2004finding}. 
The sketch is computed as $XR$, where $R \in \{-1, 0, 1\}^{p \times s}$ is a sparse matrix with 1 non-zero element per column that is $\pm 1$ with equal probability.
The nonzero column indices and nonzero values are computable using 2-universal hash functions, obviating the expensive i.i.d. sampling requirements of the FWHT. 

Importantly, computing $XR$ CST requires only $\cO(\nnz(X))$ operations. 
Also important, the rows of $XR$ preserve the sparsity of the rows of $X$.
FWHT embeddings will always be dense - even if the corresponding row has only 1 nonzero element!
These features make CST particularly suited to distributed online sketching of graphs, in which edges arrive one-by-one into working memory simultaneously on a large number of processors as highly sparse vector updates.
However, the FWHT satisfies more stringent theoretical guarantees, making it a useful baseline for comparison. 

It is import to note that both FWHT and CST can be implemented in a fully-dynamic streaming fashion on arbitrary matrices that arrive in any order and can evolve during the process of performing the embedding.
This makes FWHT and CST embeddings fully robust to changes in the underlying graph, a large advantage over the practical difficulties that face spectral methods in the streaming setting.

\section{Clustering Experiments} \label{sec:experiments}

\subsection{Experimental Setup}

In our experiments we implement an embedding-clustering pipeline similar to that of spectral clustering.
In particular, we use a sketch transform to produce an embedding into some dimension $s = \Theta(\varepsilon^{-2} \log n)$, and then further sharpen the pairwise distances between communities by applying the nonlinear dimensionality reduction tool UMAP.
We assume throughout that we know the ground truth number of communities $c$ for each graph considered, and so we use UMAP to reduce the embedding dimension from $s$ to $c$. 
We then cluster the rows of the embedding in $\mathbb{R}^{n \times c}$ using HDBSCAN.
We keep the parameters of both UMAP or HDBSCAN fixed at their default values in all experiments.

In our clustering experiments we assess the performance of our Sketch-UMAP-HDBSCAN pipeline using pairwise precision and recall~\citep{kao2017streaming} as our primary metrics. 
In particular, we determine the relationship between the parameter $\varepsilon$ - which determines the fidelity to which inner products are preserved by our embedding - with the following SBM features:
\begin{enumerate}
	\item the ratio of on-diagonal to off-diagonal entries in the probability matrix $\frac{\rho_{in}}{\rho_{out}}$;
	\item the number of communities $c$; 
	\item the number of vertices $n$.
\end{enumerate}

We test embeddings produced by both CST and FWHT, 
making the simplifying assumption that all SBMs have a probability matrix where diagonal elements equal $\rho_{in}$ and off-diagonal elements equal $\rho_{out}$.
We also assume that all blocks are equally-sized, i.e. that $\overline{\cC} = n / c$.
We apply our methods to more sophisticated SBMs that violate these assumptions in the next section.

In our experiments, we track what we refer to as the {\it maximum viable} $\varepsilon$. 
We define this to be the largest value $\varepsilon$ can achieve in a sketch embedding before either the pairwise recall or pairwise precision falls below a certain threshold. 
This maximum viable $\varepsilon$ is related to the sketch dimension of the graph matrix in an inverse squared manner~\citep{johnson1984extensions}; e.g., a reduction of epsilon by a factor of 10 leads to a 100-fold increase in embedding dimension.

In the first experiment, we vary the on- versus off-diagonal ratio of $\rho_{in}$ to $\rho_{out}$ from $\theta(10)$ to $\theta(10^3)$, fixing the number of vertices $n=4096$, the number of communities $c = 16$ and community sizes to $\cC_i = 256$ for each $i$. 
The row sums of the generated $P$ matrix are set to be 0.5, and for each sketch/parameter combination we run 10 independent trials. 
In the second two experiments we fix $\rho_{in} / \rho_{out} \sim 50$ and $\cC_i = n / c$, varying $c$ in experiment two and $n$ in experiment three.
In the second experiment, we fix $n = 4096$ while varying $c$ from 2 to 32. 
In the third experiment, we fix $c = 16$ while varying $n$ from 512 to 8096.
In all three experiments we set the thresholds for precision and recall (the \emph{metric threshold}) to 0.90, 0.95, and 0.99.
Our metric threshold dictates how large we can allow $\varepsilon$ to grow while maintaining the given tolerance. 

\subsection{Results}
\begin{figure*}[t]
\centering
\subfigure[On- vs Off-Diagonal Ratio]{
	\label{fig:diag_off_diag}
	\includegraphics[width=0.31\textwidth]{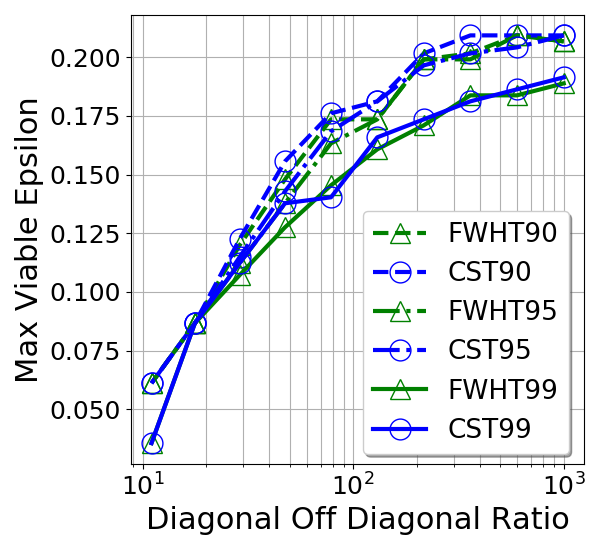}
}
\subfigure[Community Count Scaling]{
	\label{fig:num_comms}
	\includegraphics[width=0.31\textwidth]{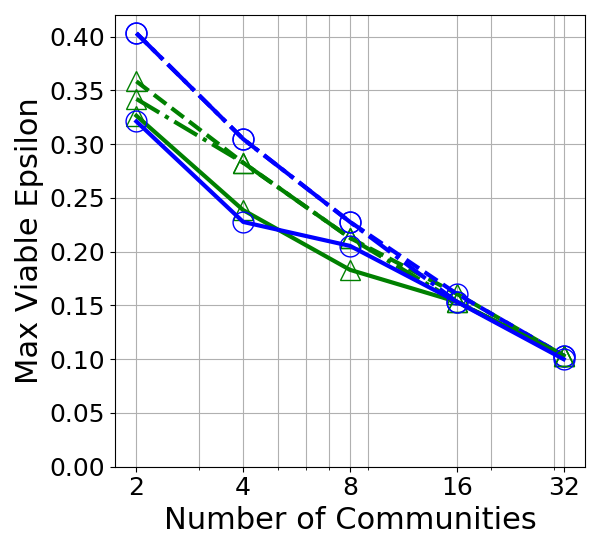}
}
\subfigure[Vertex Count Scaling]{
	\label{fig:num_verts}
	\includegraphics[width=0.31\textwidth]{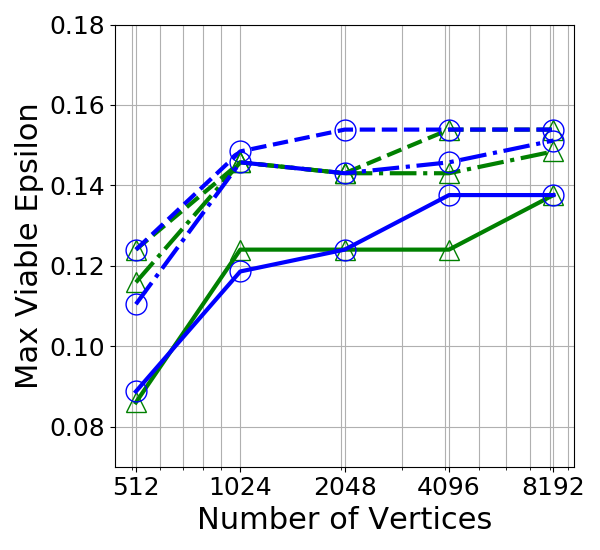}
}
\caption{
	The estimated maximum viable $\varepsilon$ as a function of SBM parameters.
	Figure~\ref{fig:diag_off_diag} plots $\varepsilon$ as a function of the ratio of the on-diagonal to off-diagonal elements in the probability matrix associated with a SBM with $16$ communities and $256$ vertices per community.
	Figure~\ref{fig:num_comms} plots $\varepsilon$ as a function of the number of communities in an SBM with $4096$ vertices and an on-diagonal to off-diagonal p matrix entry ratio of $\sim50$.
	Figure~\ref{fig:num_verts} plots $\varepsilon$  as a function of the number of vertices in an SBM with $16$ equally-sized communities and an on-diagonal to off-diagonal p matrix entry ratio of $\sim50$. 
}
\end{figure*}

Examining the first test case with results reported in Figure~\ref{fig:diag_off_diag}, we see that as the on-diagonal to off-diagonal ratio is increased, the maximum viable $\varepsilon$ increases. 
This matches our intuition; if connections between different communities are unlikely relative to those within communities, low-dimensional embeddings more accurately preserve clustering features. 
This corresponds directly to looser bounds on $\varepsilon$; if our clusters are more isolated from one another, the sketch embedding can be constructed such that it preserves geometric structure to a lesser extent. 
We also observe that the CST and FWHT perform nearly identically for all test cases, giving credibility to the use of the more distributed computationally-friendly CST in place of the more theoretically justified FWHT. 
Finally, as we decrease our metric thresholds from 0.99 to 0.90, we obtain a larger maximum viable $\varepsilon$; when we decrease our demands on cluster quality, our embedding can be of lower fidelity.

We now determine the dependence of the maximum viable $\varepsilon$ on the number of communities in our SBM. 
We expect that, as we increase the number of communities with the number of vertices fixed at $4096$, achieving desired precision/recall will become more difficult. 
Consequently, our sketch will have to map into a higher embedding dimension to sufficiently capture the geometry of the original matrix, corresponding to a lower maximum viable $\varepsilon$. 
Our results shown in Figure~\ref{fig:num_comms} confirm this prediction; as we increase the number of communities in our SBM we see that the maximum viable $\varepsilon$ grows smaller. 

Finally, we fix the number of communities in our SBM while increasing the overall number of vertices in the SBM (hence we are increasing the number of vertices per community as well). 
Figure~\ref{fig:num_verts} shows that the maximum viable $\varepsilon$ increases as the number of vertices increases
Further, as the number of vertices increases, we observe a flattening out of the maximum viable $\varepsilon$, which indicates that our methods ought to scale well as we increase the size of the graphs we are clustering, assuming that the community count remains fixed.

Across all three experiments we observe some important trends. 
First, we see that increasing our metric threshold does not drastically decrease the value of $\varepsilon$ necessary to embed our graph matrix. 
We therefore expect high performance from our methods at a relatively marginal cost. 
Further, the CST and FWHT achieve quite similar results in all test cases. 
Given that the CST is much faster and naturally implemented in a distributed setting while the FWHT satisfies more rigorous embedding properties, this is a particularly exciting result. 
Broadly, our results suggest that we ought to expect upper bounds on $\varepsilon$ to be increase as the number of communities increases, as the number of vertices per community decreases, and as the overlap between communities increases. 
On the other hand, as clusters in a graph become less disparate, our sketch embedding must capture the properties of the full graph matrix to a higher degree of fidelity.

\section{Scaling Experiments} \label{sec:scaling_experiments}

We now analyze the quality of the embedding-clustering pipeline on large benchmark graphs and test the scalability of our distributed-memory implementation of the embedding procedure. 
We utilize a selection of open-source SBMs generated as a part of the HPEC graph challenge \citep{kao2017streaming}.
Unlike our earlier experiments, these SBMs feature variable-sized communities and more complex probability matrices.
Table~\ref{tab:gc_performance} shows a selection of graph sizes, true cluster counts, and the values of $\varepsilon$ used to produce a CST embedding, as well as the pair precision (PP), pair recall (PR), and accuracy averaged over 10 independent trials.\footnote{Results using FWHT were similar.} 
In general, $\varepsilon$ was chosen to  obtain average PP and PR both $> 0.9$. 
We note, however, than as the graphs grow in size and complexity, that $\varepsilon$ also decreases. 
Figure~\ref{fig:num_comms} provides a likely explanation for this phenomenon, as community count serves as a damping factor on the maximum viable $\varepsilon$.   
Further, performant UMAP and HDBSCAN parameters most likely differ as SBMs vary in size and complexity. 

\begin{figure}[t]
\centering   
	\includegraphics[width=0.31\textwidth]{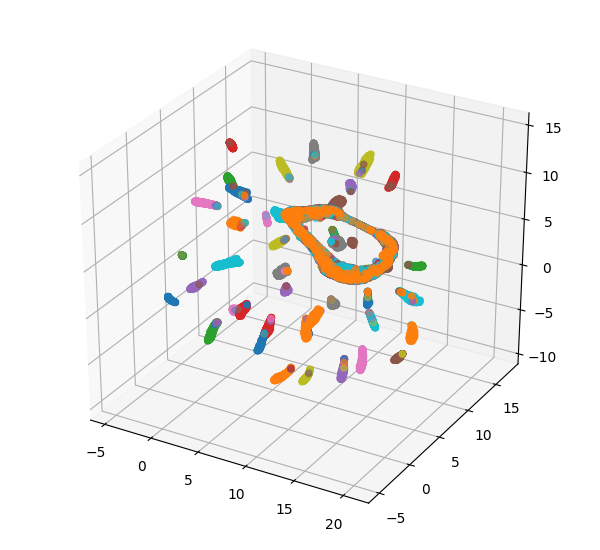}
\caption{
	A 3-dimensional visualization of the clusters produced by projecting a 50 thousand vertex SBM with 44 communities using CST with $\varepsilon=0.01$.
	Vertex embedding locations are colored according to their ground truth partition.
	Although the corresponding row of Table~\ref{tab:gc_performance} shows poor clustering results, a simple visual inspection shows that many of the true partitions separated cleanly, even in only 3 dimensions. 
	The discrepancy in analytic metrics can likely be ameliorated by recursion and/or exploration of the UMAP/HDBSCAN parameter space, which we have chosen to keep fixed.
	\label{fig:50k_clusters}
}
\end{figure}

\begin{table}[b]
\begin{center}
\begin{tabular}{|c|c|c|c|c|c|}
\hline
\textbf{$|\cV|$}&\multicolumn{5}{|c|}{\textbf{Pair Precision, Recall $> 0.9$ Parameters}} \\
\cline{2-6} 
 & \textbf{\textit{\# Partitions}} & \textbf{\textit{$\varepsilon$}} & \textbf{\textit{PP}} & \textbf{\textit{PR}} & \textbf{\textit{Accuracy}} \\
\hline
500 & 8 & $0.1$ & 0.96983 & 0.97719 & 0.986 \\
1000 & 11 & $0.1$ & 0.95991 & 0.95301 & 0.976 \\
5000 & 19 & $0.05$ & 0.97103 & 0.97395 & 0.9878 \\
20000 & 32 & $0.018$ & 0.91305 & 0.90455 & 0.9588 \\
\hline
\textbf{50000} & 44 & $0.01$ & 0.55959 & 0.12414 & 0.73773\\
\hline
\end{tabular}
\end{center}
\caption{SBM Embedding Experiments}
\label{tab:gc_performance}
\end{table}

Figure~\ref{fig:50k_clusters} shows the largest of these SBM embeddings projected down into 3 dimensions and colored according to their ground truth communities.
As we can see, even in the small dimensional space, the embedding manages to separate most of the clusters, some completely, others mixed into clusters of 2 or more communities. 
This suggests that our method, like most others at scale, will likely be best applied by hierarchically partitioning and refining subsets of vertices.

\subsection{Distributed Sketching}

A sketch embedding of dimension $s$ on a square graph matrix $X \in \mathbb{R}^{n \times n}$ of $\cG$ is performed as follows:
\begin{enumerate}
  \item Choose a desired precision $\varepsilon \in (0, 1)$.
  \item Sample a sketch operator $S \in \mathbb{R}^{n \times s}$, with $s = O(\varepsilon^{-2})$.
  \item Compute sketch $XS$.
  \item Identify with each vertex $x$ the row vector $(XS)_{x, :}$.
\end{enumerate}

We distribute this procedure as follows.
Assume a universe of processors $\cP$, and further assume some arbitrary balanced partitioning of vertices to processors $f : \cV \rightarrow \cP$.\footnote{We use simple round-robin assignment in our experiments.}
We will abuse notation and refer to the set of vertices assigned to $P \in \cP$ by $f$ as $f^{-1}(P)$ .
Let $\boldsymbol{\sigma}$ be an arbitrary stream of edge updates defining $\cE$, partitioned such that each $P \in \cP$ receives the substream $\boldsymbol{\sigma}_P$.  
Each $P \in \cP$ maintains a sketch vector in $\mathbb{R}^s$ corresponding to $(XS)_{x. :}$ for each $x \in f^{-1}(P)$.
On reading an edge $uv \in \boldsymbol{\sigma}_P$, $P$ sends $uv$ to $f(u)$ and $vu$ to $f(v)$.
Upon receiving an edge $xy$, $f(x) = P$, processor $P$ updates $(XS)_{x,:}$ appropriately.
After having read over $\boldsymbol{\sigma}$ and cleared their communication buffers, $\cP$ has $XS$ stored in distributed memory.

We examine the scaling limits of the embedding procedure by implementing our distributed sketches using the C++/MPI communication library YGM~\citep{priest2019you}
and applying them to very large SBMs generated with GraphChallenge 2017 parameters, which we fit with constrained regression.
We use
\begin{align*}
c &= 0.95*n^{-0.36}, 
&\rho_{in} &\sim 16.75 n^{-0.59},
\\  
\overline{\cC} &= .95n^{0.64}, 
&\rho_{out} &\sim -1.02 n^{-0.59},
\\
Var(\cC) &= .32n^{0.64},
\end{align*}
where $c$ is the number of communities, $\overline{\cC}, Var(\cC)$ are the parameters used for sampling community sizes, and $\rho_{in}, \rho_{out}$ are internal/external edge density parameters used in the SBM generation.  
All of our distributed experiments were performed on a cluster of Intel Xeon E5-2695 processors each featuring 36 cores.

\begin{figure*}
\centering     
\subfigure[Fixed Resource Scaling]{
	\label{fig:fixed_scaling}
	\includegraphics[width=0.31\textwidth]{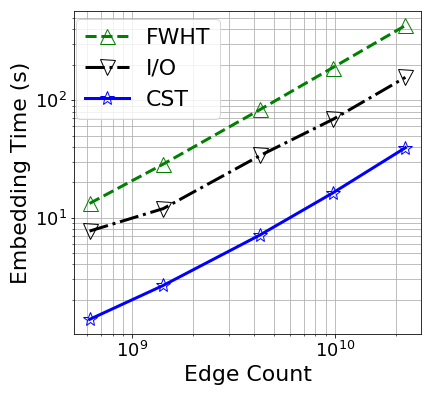}
}
\subfigure[Strong Scaling]{
	\label{fig:strong_scaling}
	\includegraphics[width=0.31\textwidth]{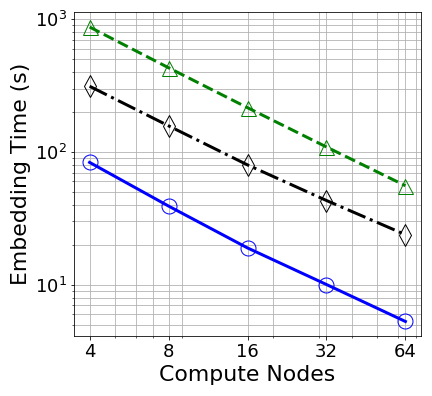}
}
\subfigure[Embedding Dimension Scaling]{
	\label{fig:dimension_scaling}
	\includegraphics[width=0.31\textwidth]{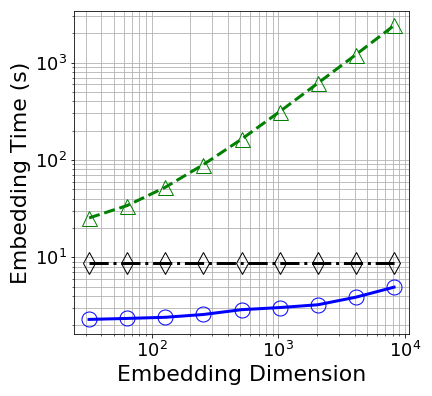}
}
\caption{
	The distributed memory scaling of applying CST and FWHT to SBMs of various sizes. We also plot the total time spent reading the graphs from file for comparison.
	Figure~\ref{fig:fixed_scaling} plots the wall time for a set of 4 compute nodes as the edge count of the graph to be embedded into 128 dimensions increases from $\sim 500$ million to $\sim 22$ billion.
	Figure~\ref{fig:strong_scaling} plots the wall time for embedding a fixed-size graph with 200 million vertices ($\sim 22$ billion edges) into 128 dimensions as the number of compute nodes increases.
	Both Figures~\ref{fig:fixed_scaling} and ~\ref{fig:strong_scaling} feature roughly linear scaling, as desired. 
	Figure~\ref{fig:dimension_scaling} plots the wall time for four compute nodes to embed a 20 million vertex graph as the embedding dimension increases, i.e. as $\varepsilon$ decreases. 
}
\end{figure*}

Figure~\ref{fig:fixed_scaling} shows the wall time scaling of our codes with a fixed number of processors as graph size increases.
Figure~\ref{fig:strong_scaling} shows the wall time where instead the graph to be embedded is fixed and we increase the number of compute nodes.
Finally, Figure~\ref{fig:dimension_scaling} shows the scaling where only the embedding dimension increases - i.e. $\varepsilon$ decreases.
These scaling studies reinforce our assertion of the fitness of CST for generating high quality low-dimensional embeddings for clustering applications, and highlight the weaknesses of the more disciplined but also more cumbersome FWHT.
In particular, we find that our implementation scales at a rate no worse than reading the graph into memory.

\section{Conclusions and Future Work}

We have demonstrated a scalable vertex embedding procedure using linear sketches, and have validated its utility and scalability on partitioning SBMs.
We have shown that this approach provides an algorithmic workflow similar to spectral clustering at a fraction of the cost, and with the benefit of much higher scalability. 

A major limiting factor to our analysis at scale is that we have limited our scope to local-parallelism-only implementations of clustering algorithms.
However, we have shown that our embedding algorithm features excellent scalability in distributed memory, and is able to embed graphs with tens of billions of edges in seconds on modest hardware.
Further, our sensitivity experiments suggest that for a fixed desired precision, we find embeddings that yield clusters whose quality has at most a small dependence upon $n$, although the dependance upon the number of true communities and their size variance warrants further investigation.
In future work we will demonstrate the scalability of the full clustering pipeline to distributed memory data scales by the introduction of novel distributed clustering algorithms. 

It is further important to recall that, though SBMs feature convenient analytical properties, they do not reflect many properties of real graphs found in applications.
Indeed, one of the largest challenges associated with distributed graph algorithms is managing the communication and computation bottlenecks introduced by the presence of high-degree vertices in scale-free graphs.
Degree-Corrected Stochastic Block Models (DCSBMs) generalize SBMs with power-law degree distributions so as to more accurately simulate this feature.
Detection of large dense regions (degree-corrected quasi-cliques) injected into real-world graphs would also be an important validation step. 
We will augment our algorithms in future work to manage high-degree vertices via \emph{vertex delegation} \citep{pearce2014faster} and sparse sketch storage.

Vertex embedding has many applications within graph machine learning beyond clustering.
For example, recent vertex representation learning efforts such as node2vec \citep{grover2016node2vec} utilize deep neural networks to construct an embedding of vertices into low-dimensional latent space.
Scaling to massive graph sizes, however, remains challenging.
We believe that linear sketch-based embedding such as what we have proposed could significantly scale such nonlinear embeddings at a negligible cost to the representation quality.

 \section*{Acknowledgements}
 
 The authors would like to thank Van Emden Henson for helpful comments and discussion.
 
\newpage
\bibliographystyle{arxiv}
\bibliography{bibliography.bib}

\begin{thebibliography}{41}
\providecommand{\natexlab}[1]{#1}
\providecommand{\url}[1]{\texttt{#1}}
\expandafter\ifx\csname urlstyle\endcsname\relax
  \providecommand{\doi}[1]{doi: #1}\else
  \providecommand{\doi}{doi: \begingroup \urlstyle{rm}\Url}\fi

\bibitem[Ahn et~al.(2012{\natexlab{a}})Ahn, Guha, and
  McGregor]{ahn2012analyzing}
Kook~Jin Ahn, Sudipto Guha, and Andrew McGregor.
\newblock Analyzing graph structure via linear measurements.
\newblock In \emph{Proceedings of the twenty-third annual ACM-SIAM symposium on
  Discrete Algorithms}, pp.\  459--467. SIAM, 2012{\natexlab{a}}.

\bibitem[Ahn et~al.(2012{\natexlab{b}})Ahn, Guha, and McGregor]{ahn2012graph}
Kook~Jin Ahn, Sudipto Guha, and Andrew McGregor.
\newblock Graph sketches: sparsification, spanners, and subgraphs.
\newblock In \emph{Proceedings of the 31st ACM SIGMOD-SIGACT-SIGAI symposium on
  Principles of Database Systems}, pp.\  5--14. ACM, 2012{\natexlab{b}}.

\bibitem[Ahn et~al.(2013)Ahn, Guha, and McGregor]{ahn2013spectral}
Kook~Jin Ahn, Sudipto Guha, and Andrew McGregor.
\newblock Spectral sparsification in dynamic graph streams.
\newblock In \emph{Approximation, Randomization, and Combinatorial
  Optimization. Algorithms and Techniques}, pp.\  1--10. Springer, 2013.

\bibitem[Ahn et~al.(2018)Ahn, Lee, and Suh]{ahn2018hypergraph}
Kwangjun Ahn, Kangwook Lee, and Changho Suh.
\newblock Hypergraph spectral clustering in the weighted stochastic block
  model.
\newblock \emph{IEEE Journal of Selected Topics in Signal Processing},
  12\penalty0 (5):\penalty0 959--974, 2018.

\bibitem[Ailon \& Chazelle(2009)Ailon and Chazelle]{ailon2009fast}
Nir Ailon and Bernard Chazelle.
\newblock The fast {J}ohnson--{L}indenstrauss transform and approximate nearest
  neighbors.
\newblock \emph{SIAM Journal on computing}, 39\penalty0 (1):\penalty0 302--322,
  2009.

\bibitem[Andoni et~al.(2011)Andoni, Krauthgamer, and Onak]{andoni2011streaming}
Alexandr Andoni, Robert Krauthgamer, and Krzysztof Onak.
\newblock Streaming algorithms via precision sampling.
\newblock In \emph{Foundations of Computer Science (FOCS), 2011 IEEE 52nd
  Annual Symposium on}, pp.\  363--372. IEEE, 2011.

\bibitem[Baker et~al.(2009)Baker, Hetmaniuk, Lehoucq, and Thornquist]{Anasazi}
C.~G. Baker, U.~L. Hetmaniuk, R.~B. Lehoucq, and H.~K. Thornquist.
\newblock Anasazi software for the numerical solution of large-scale eigenvalue
  problems.
\newblock \emph{ACM Trans. Math. Softw.}, 36\penalty0 (3), July 2009.
\newblock ISSN 0098-3500.
\newblock \doi{10.1145/1527286.1527287}.
\newblock URL \url{https://doi.org/10.1145/1527286.1527287}.

\bibitem[Bulu{\c{c}} \& Gilbert(2011)Bulu{\c{c}} and
  Gilbert]{bulucc2011combinatorial}
Ayd{\i}n Bulu{\c{c}} and John~R Gilbert.
\newblock The combinatorial {BLAS}: Design, implementation, and applications.
\newblock \emph{The International Journal of High Performance Computing
  Applications}, 25\penalty0 (4):\penalty0 496--509, 2011.

\bibitem[Charikar et~al.(2004)Charikar, Chen, and
  Farach-Colton]{charikar2004finding}
Moses Charikar, Kevin Chen, and Martin Farach-Colton.
\newblock Finding frequent items in data streams.
\newblock \emph{Theoretical Computer Science}, 312\penalty0 (1):\penalty0
  3--15, 2004.

\bibitem[Clarkson \& Woodruff(2017)Clarkson and Woodruff]{clarkson2017low}
Kenneth~L Clarkson and David~P Woodruff.
\newblock Low-rank approximation and regression in input sparsity time.
\newblock \emph{Journal of the ACM (JACM)}, 63\penalty0 (6):\penalty0 1--45,
  2017.

\bibitem[Fowlkes et~al.(2004)Fowlkes, Belongie, Chung, and
  Malik]{fowlkes2004spectral}
Charless Fowlkes, Serge Belongie, Fan Chung, and Jitendra Malik.
\newblock Spectral grouping using the {N}ystr\"{o}m method.
\newblock \emph{IEEE transactions on pattern analysis and machine
  intelligence}, 26\penalty0 (2):\penalty0 214--225, 2004.

\bibitem[{Ghosh} et~al.(2019){Ghosh}, {Halappanavar}, {Tumeo}, and
  {Kalyanarainan}]{Ghosh18GC}
S.~{Ghosh}, M.~{Halappanavar}, A.~{Tumeo}, and A.~{Kalyanarainan}.
\newblock Scaling and quality of modularity optimization methods for graph
  clustering.
\newblock In \emph{2019 IEEE High Performance Extreme Computing Conference
  (HPEC)}, pp.\  1--6, 2019.

\bibitem[Gilbert et~al.(2012)Gilbert, Park, and Wakin]{gilbert2012sketched}
Anna~C Gilbert, Jae~Young Park, and Michael~B Wakin.
\newblock Sketched {SVD}: Recovering spectral features from compressive
  measurements.
\newblock \emph{arXiv preprint arXiv:1211.0361}, 2012.

\bibitem[Gittens et~al.(2013)Gittens, Kambadur, and
  Boutsidis]{gittens2013approximate}
Alex Gittens, Prabhanjan Kambadur, and Christos Boutsidis.
\newblock Approximate spectral clustering via randomized sketching.
\newblock \emph{Ebay/IBM Research Technical Report}, 2013.

\bibitem[Grover \& Leskovec(2016)Grover and Leskovec]{grover2016node2vec}
Aditya Grover and Jure Leskovec.
\newblock node2vec: Scalable feature learning for networks.
\newblock In \emph{Proceedings of the 22nd ACM SIGKDD international conference
  on Knowledge discovery and data mining}, pp.\  855--864, 2016.

\bibitem[Hahsler et~al.(2019)Hahsler, Piekenbrock, and
  Doran]{hahsler2019dbscan}
Michael Hahsler, Matthew Piekenbrock, and Derek Doran.
\newblock dbscan: Fast density-based clustering with r.
\newblock \emph{Journal of Statistical Software}, 91\penalty0 (1):\penalty0
  1--30, 2019.

\bibitem[{Halappanavar} et~al.(2017){Halappanavar}, {Lu}, {Kalyanaraman}, and
  {Tumeo}]{Halappanavar17GC}
M.~{Halappanavar}, H.~{Lu}, A.~{Kalyanaraman}, and A.~{Tumeo}.
\newblock Scalable static and dynamic community detection using grappolo.
\newblock In \emph{2017 IEEE High Performance Extreme Computing Conference
  (HPEC)}, pp.\  1--6, 2017.

\bibitem[Halko et~al.(2011)Halko, Martinsson, and Tropp]{halko2011finding}
Nathan Halko, Per-Gunnar Martinsson, and Joel~A Tropp.
\newblock Finding structure with randomness: Probabilistic algorithms for
  constructing approximate matrix decompositions.
\newblock \emph{SIAM review}, 53\penalty0 (2):\penalty0 217--288, 2011.

\bibitem[Hernandez et~al.(2005)Hernandez, Roman, and Vidal]{Hernandez:2005:SSF}
Vicente Hernandez, Jose~E. Roman, and Vicente Vidal.
\newblock {SLEPc}: A scalable and flexible toolkit for the solution of
  eigenvalue problems.
\newblock \emph{{ACM} Trans. Math. Software}, 31\penalty0 (3):\penalty0
  351--362, 2005.

\bibitem[Johnson \& Lindenstrauss(1984)Johnson and
  Lindenstrauss]{johnson1984extensions}
William~B Johnson and Joram Lindenstrauss.
\newblock Extensions of {L}ipschitz mappings into a {H}ilbert space.
\newblock \emph{Contemporary mathematics}, 26\penalty0 (189-206):\penalty0 1,
  1984.

\bibitem[Jowhari et~al.(2011)Jowhari, Sa{\u{g}}lam, and
  Tardos]{jowhari2011tight}
Hossein Jowhari, Mert Sa{\u{g}}lam, and G{\'a}bor Tardos.
\newblock Tight bounds for lp samplers, finding duplicates in streams, and
  related problems.
\newblock In \emph{Proceedings of the thirtieth ACM SIGMOD-SIGACT-SIGART
  symposium on Principles of database systems}, pp.\  49--58. ACM, 2011.

\bibitem[Kao et~al.(2017)Kao, Gadepally, Hurley, Jones, Kepner, Mohindra,
  Monticciolo, Reuther, Samsi, Song, et~al.]{kao2017streaming}
Edward Kao, Vijay Gadepally, Michael Hurley, Michael Jones, Jeremy Kepner,
  Sanjeev Mohindra, Paul Monticciolo, Albert Reuther, Siddharth Samsi, William
  Song, et~al.
\newblock Streaming graph challenge: Stochastic block partition.
\newblock In \emph{2017 IEEE High Performance Extreme Computing Conference
  (HPEC)}, pp.\  1--12. IEEE, 2017.

\bibitem[Kapralov et~al.(2017)Kapralov, Lee, Musco, Musco, and
  Sidford]{kapralov2017single}
Michael Kapralov, Yin~Tat Lee, CN~Musco, CP~Musco, and Aaron Sidford.
\newblock Single pass spectral sparsification in dynamic streams.
\newblock \emph{SIAM Journal on Computing}, 46\penalty0 (1):\penalty0 456--477,
  2017.

\bibitem[Knyazev(2018)]{knyazev2018spectral}
Andrew Knyazev.
\newblock On spectral partitioning of signed graphs.
\newblock In \emph{2018 Proceedings of the Seventh SIAM Workshop on
  Combinatorial Scientific Computing}, pp.\  11--22. SIAM, 2018.

\bibitem[Li et~al.(2011)Li, Lian, Kwok, and Lu]{li2011time}
Mu~Li, Xiao-Chen Lian, James~T Kwok, and Bao-Liang Lu.
\newblock Time and space efficient spectral clustering via column sampling.
\newblock In \emph{CVPR 2011}, pp.\  2297--2304. IEEE, 2011.

\bibitem[{Liu} et~al.(2019){Liu}, {Firoz}, {Zalewski}, {Halappanavar},
  {Barker}, {Lumsdaine}, and {Gebremedhin}]{Liu18GC}
X.~{Liu}, J.~S. {Firoz}, M.~{Zalewski}, M.~{Halappanavar}, K.~J. {Barker},
  A.~{Lumsdaine}, and A.~H. {Gebremedhin}.
\newblock Distributed direction-optimizing label propagation for community
  detection.
\newblock In \emph{2019 IEEE High Performance Extreme Computing Conference
  (HPEC)}, pp.\  1--6, 2019.

\bibitem[Lloyd(1982)]{lloyd1982least}
Stuart Lloyd.
\newblock Least squares quantization in {PCM}.
\newblock \emph{IEEE transactions on information theory}, 28\penalty0
  (2):\penalty0 129--137, 1982.

\bibitem[Martin et~al.(2017)Martin, Loukas, and Vandergheynst]{martin2017fast}
Lionel Martin, Andreas Loukas, and Pierre Vandergheynst.
\newblock Fast approximate spectral clustering for dynamic networks, 2017.

\bibitem[McInnes et~al.(2017)McInnes, Healy, and Astels]{mcinnes2017hdbscan}
Leland McInnes, John Healy, and Steve Astels.
\newblock hdbscan: Hierarchical density based clustering.
\newblock \emph{Journal of Open Source Software}, 2\penalty0 (11):\penalty0
  205, 2017.

\bibitem[McInnes et~al.(2018)McInnes, Healy, and Melville]{mcinnes2018umap}
Leland McInnes, John Healy, and James Melville.
\newblock {UMAP}: Uniform manifold approximation and projection for dimension
  reduction.
\newblock \emph{arXiv preprint arXiv:1802.03426}, 2018.

\bibitem[Ning et~al.(2007)Ning, Xu, Chi, Gong, and Huang]{Ning2007}
Huazhong Ning, Wei Xu, Yun Chi, Yihong Gong, and Thomas~S. Huang.
\newblock Incremental spectral clustering with application to monitoring of
  evolving blog communities.
\newblock In \emph{Proceedings of the Seventh {SIAM} International Conference
  on Data Mining, April 26-28, 2007, Minneapolis, Minnesota, {USA}}, pp.\
  261--272. {SIAM}, 2007.
\newblock \doi{10.1137/1.9781611972771.24}.
\newblock URL \url{https://doi.org/10.1137/1.9781611972771.24}.

\bibitem[Papadimitriou et~al.(2000)Papadimitriou, Raghavan, Tamaki, and
  Vempala]{papadimitriou2000latent}
Christos~H Papadimitriou, Prabhakar Raghavan, Hisao Tamaki, and Santosh
  Vempala.
\newblock Latent semantic indexing: A probabilistic analysis.
\newblock \emph{Journal of Computer and System Sciences}, 61\penalty0
  (2):\penalty0 217--235, 2000.

\bibitem[Pearce et~al.(2014)Pearce, Gokhale, and Amato]{pearce2014faster}
Roger Pearce, Maya Gokhale, and Nancy~M Amato.
\newblock Faster parallel traversal of scale free graphs at extreme scale with
  vertex delegates.
\newblock In \emph{High Performance Computing, Networking, Storage and
  Analysis, SC14: International Conference for}, pp.\  549--559. IEEE, 2014.

\bibitem[Priest et~al.(2019)Priest, Steil, Pearce, and Sanders]{priest2019you}
Benjamin Priest, Trevor Steil, Roger Pearce, and Geoff Sanders.
\newblock You've {G}ot {M}ail: Building missing asynchronous communication
  primitives.
\newblock In \emph{Proceedings of the 2019 International Conference on
  Supercomputing}, pp.\ ~8. ACM, 2019.

\bibitem[{Que} et~al.(2015){Que}, {Checconi}, {Petrini}, and
  {Gunnels}]{Que15Louvain}
X.~{Que}, F.~{Checconi}, F.~{Petrini}, and J.~A. {Gunnels}.
\newblock Scalable community detection with the {L}ouvain algorithm.
\newblock In \emph{2015 IEEE International Parallel and Distributed Processing
  Symposium}, pp.\  28--37, 2015.

\bibitem[Rokhlin \& Tygert(2008)Rokhlin and Tygert]{rokhlin2008fast}
Vladimir Rokhlin and Mark Tygert.
\newblock A fast randomized algorithm for overdetermined linear least-squares
  regression.
\newblock \emph{PNAS}, 105\penalty0 (36):\penalty0 13212--13217, 2008.

\bibitem[Traganitis et~al.(2015)Traganitis, Slavakis, and
  Giannakis]{traganitis2015sketch}
Panagiotis~A Traganitis, Konstantinos Slavakis, and Georgios~B Giannakis.
\newblock Sketch and validate for big data clustering.
\newblock \emph{IEEE Journal of Selected Topics in Signal Processing},
  9\penalty0 (4):\penalty0 678--690, 2015.

\bibitem[Van~Lierde et~al.(2018)Van~Lierde, Chow, and Delvenne]{VanLierde}
Hadrien Van~Lierde, Tommy W~S Chow, and Jean-Charles Delvenne.
\newblock {Spectral clustering algorithms for the detection of clusters in
  block-cyclic and block-acyclic graphs}.
\newblock \emph{Journal of Complex Networks}, 7\penalty0 (1):\penalty0 1--53,
  05 2018.
\newblock ISSN 2051-1329.
\newblock \doi{10.1093/comnet/cny011}.
\newblock URL \url{https://doi.org/10.1093/comnet/cny011}.

\bibitem[Von~Luxburg(2007)]{von2007tutorial}
Ulrike Von~Luxburg.
\newblock A tutorial on spectral clustering.
\newblock \emph{Statistics and computing}, 17\penalty0 (4):\penalty0 395--416,
  2007.

\bibitem[Woodruff et~al.(2014)]{woodruff2014sketching}
David~P Woodruff et~al.
\newblock Sketching as a tool for numerical linear algebra.
\newblock \emph{Foundations and Trends{\textregistered} in Theoretical Computer
  Science}, 10\penalty0 (1--2):\penalty0 1--157, 2014.

\bibitem[Yang et~al.(2020)Yang, Liu, Dobriban, and Woodruff]{yang2020reduce}
Fan Yang, Sifan Liu, Edgar Dobriban, and David~P Woodruff.
\newblock How to reduce dimension with {PCA} and random projections?
\newblock \emph{arXiv preprint arXiv:2005.00511}, 2020.

\end{thebibliography}

\end{document}